\documentclass[runningheads]{llncs}
\usepackage[T1]{fontenc}
\usepackage{graphicx,verbatim}
\usepackage{xurl}
\usepackage{hyperref}
\usepackage{amsmath}
\usepackage{booktabs}
\usepackage{bbding}
\usepackage{tabularx}
\usepackage{amsfonts} %
\begin{document}
\title{CardioTrace: Motion-Aware Cine Synthesis via Coupled Tracing and Segmentation}
\titlerunning{CardioTrace}
%
\author{Zhi Chen\inst{1}\and
Le Zhang\inst{1}}
\authorrunning{Z. Chen, L. Zhang}
%
\institute{School of Engineering, College of Engineering and Physical Sciences, University of Birmingham, Birmingham, UK\\
\email{zxc561@student.bham.ac.uk}}


  
\maketitle              

\begin{abstract}
Cardiac cine MRI provides time-resolved visualization of cardiac motion and is essential for evaluating ventricular function and anatomy. In routine clinical acquisition, however, cine sequences are frequently temporally sparse or degraded by arrhythmia, motion, and scan-time constraints, which can hinder reliable quantification and downstream analysis. We present CardioTrace, a motion-aware cine synthesis framework that generates a full cardiac cine sequence from limited input while explicitly modeling and tracing cardiac motion. CardioTrace couples two complementary pathways: a pixel-translation branch that directly predicts cine frames in a residual formulation, and a flow-field branch that estimates time-resolved deformation fields from a reference phase to provide interpretable motion correspondences across frames. To further preserve anatomical plausibility, a segmentation branch supervises clinically meaningful cardiac structures at key phases, and a cross-branch consistency objective encourages the synthesized appearance dynamics to follow the explicit motion guidance, mitigating temporal drift and unrealistic shape transitions over long sequences. The framework also supports prompt-style conditioning to incorporate auxiliary patient or acquisition information, improving robustness across heterogeneous multi-center data. Experiments on public cardiac cine MRI datasets demonstrate that CardioTrace improves perceptual quality and temporal coherence over representative GAN- and diffusion-based baselines, while better preserving ventricular morphology throughout the cardiac cycle.
\textit{Our codes are available after review}.
\keywords{Cardiac MRI  \and Image synthesis \and Temporal consistency.}

\end{abstract}

\section{Introduction}

Cardiac cine MRI is an imaging technique used to assess cardiac function and anatomical structures by capturing a series of magnetic resonance images.
It provides time-resolved visualization of ventricular motion and enabling quantitative measurements such as volumes and ejection fraction.
However, in practical applications, cine sequences are often temporally sparse or affected by motion, arrhythmias, and acquisition constraints, which pose challenges to clinical diagnosis.
Therefore, learning-based methods are needed to synthesize temporally consistent sequences from limited or imperfect data, such as~\cite{huang2024subspaceimplicitneuralrepresentations},
in order to support accelerated imaging and ensure reliable downstream analysis. Recent advances in deep generative models, including GAN-based~\cite{vukadinovic2023gancmri} and diffusion-based approaches~\cite{qiu2024spatiotemporaldiffusionmodelpaired}, have improved image realism across translation tasks.
Despite promising visual quality, existing learning-based methods often suffer from temporal inconsistency, subtle anatomical distortions, and limited interpretability~\cite{yoo2020generative}. Clinically reliable cine synthesis therefore requires anatomically consistent results with explicit and transparent motion modeling.

Below, we highlight the key challenge that needs be addressed
for cardiac cine MRI synthesis: 
\textit{How to effectively use cardiac motion to supervise image generation and establish cross-domain constraints?}
GANcMRI~\cite{vukadinovic2023gancmri} explores cardiac cine sequence generation by modeling temporal trajectories in the latent space, and reports high pixel-level similarity to ground-truth sequences. 
However, due to the lack of explicit spatial or anatomical constraints, the synthesized sequences may exhibit implausible deformations and motion-related artifacts.
These limitations indicate the need to introduce alternative approaches that can explicitly model cardiac motion while maintaining long-range temporal consistency.
Motion- or registration-based
consistency is often introduced to couple deformation
estimation with structural constraints~\cite{zheng2024deformation}. By enforcing correspondence or temporal consistency across frames,
such strategies provide an additional form of weak supervision that
does not rely on dense annotations~\cite{luo2025tracetrans}. Some works like TraceTrans~\cite{luo2025tracetrans} enable post-operative prediction from pre-operative inputs by explicitly modeling spatial deformations and using them to synthesize anatomically consistent translated images. However, when faced with longer cine sequences, these approaches often suffer from error accumulation: small registration inaccuracies can compound over time, leading to temporal drift and gradually increasing anatomical misalignment across frames.

Therefore, we propose CardioTrace, a unified framework that integrates cardiac cine synthesis with explicit and traceable motion modeling. 
By jointly generating cine frames together with time-resolved motion representations, CardioTrace establishes interpretable correspondences across frames, thereby improving long-range temporal consistency and reducing implausible shape transitions. 
The explicit motion representation further facilitates model inspection and downstream analysis, and can optionally incorporate patient-specific information to enhance robustness and generalization across heterogeneous data sources.

\textbf{Our contributions:}
\textbf{(1)} We propose a unified learning framework that jointly models deformation-field-based motion tracing, cardiac segmentation for anatomical supervision, and direct cine frame prediction, enabling complementary objectives to reinforce temporal consistency and structural fidelity. 
\textbf{(2)} \textbf{Network design with prompt embedding:} We introduce an architecture tailored for motion-traceable cine synthesis, and incorporate prompt embeddings to inject auxiliary information into the network, improving controllability and robustness across heterogeneous cardiac MRI data. 
\textbf{(3)} Comprehensive experiments on multi-center cardiac cine MRI demonstrate improved image quality and temporal coherence over competitive baselines, while better preserving clinically meaningful cardiac structures across the cardiac cycle.

\begin{figure}[t]
    \centering
    \includegraphics[width=\textwidth]{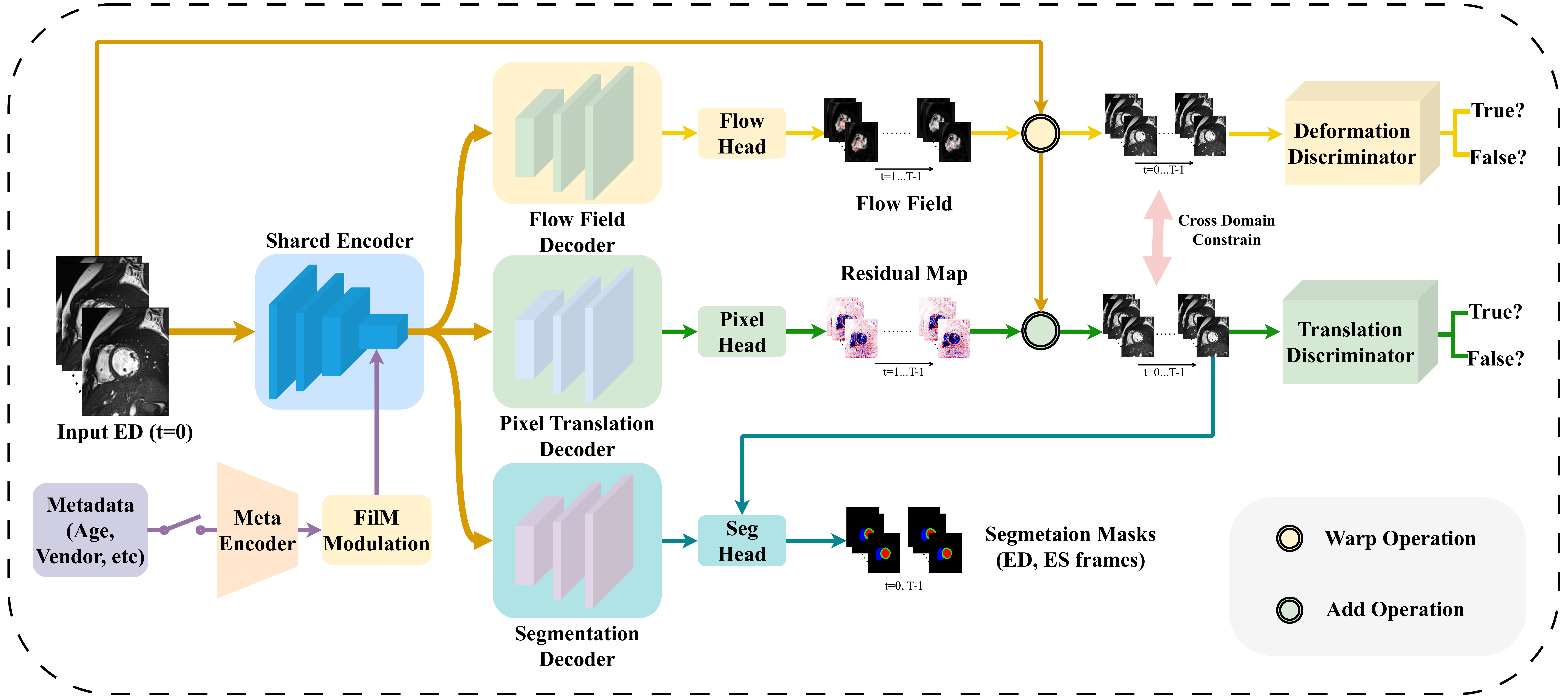}
    \caption{Overview of the CardioTrace framework. An encoder extracts features from the ED frame, which are processed by three decoders: the Pixel Translation Branch generates the sequence prediction, and the Flow Field Branch predicts the deformation field to ensure structural correspondence among the whole sequence, while discriminators optimize translation fidelity and anatomical consistency. The Segmentation Branch does segmentation at end-diastolic (ED) and end-systolic (ES) frames.}
    \label{framework}
\end{figure}

\section{Methodology}
\subsection{Problem Formulation}
Given an end-diastolic (ED) cardiac MR volume $I_0 \in \mathbb{R}^{1\times Z\times H\times W}$, our goal is to synthesize a temporally coherent cine sequence
$I=\{I_t\}_{t=0}^{T-1}$ that captures cardiac motion and appearance evolution, while preserving anatomical correctness across clinically meaningful phases.
Beyond image synthesis, we explicitly model cardiac motion and leverage sparse anatomical supervision available.

To this end, we formulate cardiac cine synthesis as a joint problem of motion estimation, pixel-wise appearance translation, and anatomical supervision.
The model predicts a set of absolute displacement fields
$\{u_{0\rightarrow t}\}_{t=1}^{T-1}$, where each
$u_{0\rightarrow t}\in\mathbb{R}^{3\times Z\times H\times W}$ maps voxels from the ED reference frame to time $t$.
These flows define a motion-driven transformation of the ED volume, yielding a motion-guided cine representation
$\hat{I}^{\text{flow}}=\{\hat{I}^{\text{flow}}_t\}_{t=0}^{T-1}$ by warping $I_0$, with $\hat{I}^{\text{flow}}_0 = I_0$.

In parallel, a Pixel Translation Branch directly synthesizes the full cine sequence
$\hat{I}^{\text{pix}}=\{\hat{I}^{\text{pix}}_t\}_{t=0}^{T-1}$ using a residual formulation relative to the ED frame.
The Flow Field Branch provides an explicit, motion-traceable representation of cardiac dynamics, while the Pixel Translation Branch focuses on appearance translation.
To couple motion, appearance, and anatomy, segmentation labels available at ED and end-systolic (ES) phases are used to supervise anatomical structures.
Furthermore, the motion-guided cine representation serves as a teacher signal for the Pixel Translation Branch, encouraging coherent temporal dynamics and reducing visually implausible frame-to-frame variations.

\subsection{Model Architecture}
The overall architecture of CardioTrace is illustrated in Fig.~\ref{framework}.
Our architecture adopts a shared-encoder, multi-branch design that jointly supports motion estimation, pixel-wise image translation, and anatomical supervision within a unified framework.
A shared encoder $\mathbf{E}$ processes the ED volume and extracts multi-scale features.
It follows a standard 3D U-Net~\cite{cciccek20163d} design with four downsampling stages, progressively increasing channel dimensionality from 64 to 512 at the bottleneck.
Each stage consists of two $3\times3\times3$ convolutions followed by Group Normalization and ReLU activation.

On top of the shared encoder, three decoder branches with symmetric upsampling and skip connections are employed.
The Flow Field Branch predicts features used to regress the displacement fields $\{u_{0\rightarrow t}\}$, which are constrained by a scaled \texttt{tanh} activation to limit maximum voxel displacement.
The Pixel Translation Branch outputs features for predicting per-frame residual maps, which are added to the ED volume to synthesize the cine sequence $\hat{I}^{\text{pix}}$.
The Segmentation Branch produces logits for three anatomical classes—the left ventricle (LV), right ventricle (RV), and myocardium (MYO)—and is applied at the ED and ES frames to provide anatomical supervision.

To improve robustness across multi-center data, optional patient meta-data are incorporated as prompt-style conditioning signals.
Categorical attributes (e.g., vendor, pathology, sex) are embedded and concatenated with normalized numerical attributes.
The resulting meta-feature vector modulates the encoder bottleneck via feature-wise linear modulation (FiLM)~\cite{perez2018film}.

Adversarial learning is employed using two 3D PatchGAN~\cite{isola2017image} discriminators with hinge loss.
The discriminator operating on motion-guided frames enforces realism under physically plausible deformations, while the discriminator conditioned on the ED reference takes concatenated pairs $[I_0, \hat{I}^{\text{pix}}_t]$ to stabilize pixel-wise translation and preserve consistency with the reference anatomy.

\subsection{Overall Loss Function}
CardioTrace is trained with a single generator objective and a single discriminator objective.
Let the ground-truth cine sequence be $I=\{I_t\}_{t=0}^{T-1}$ with $I_t\in\mathbb{R}^{1\times Z\times H\times W}$, and the ED input be $I_0$.
The Flow Field Branch outputs absolute flows $\{u_{0\rightarrow t}\}_{t=1}^{T-1}$ and produces a deformation-based (flow-guided) sequence
$\hat{I}^{\text{flow}}=\{\hat{I}^{\text{flow}}_t\}_{t=0}^{T-1}$ by warping $I_0$.
The Pixel Translation Branch produces a translation-based sequence
$\hat{I}^{\text{pix}}=\{\hat{I}^{\text{pix}}_t\}_{t=0}^{T-1}$ via residual prediction on $I_0$.

\textbf{Reconstruction loss $\mathcal{L}_{\text{recon}}$.}
We enforce frame-wise fidelity to the ground-truth sequence for both branches over $t=1,\dots,T-1$:
\begin{equation}
\mathcal{L}_{\text{recon}}
=
\lambda_{\text{flow-img}}\mathcal{L}_{\text{def}}
+
\lambda_{\text{pix-img}}\mathcal{L}_{\text{trans}},
\end{equation}
where
\begin{equation}
\mathcal{L}_{\text{def}}
=
\frac{1}{T-1}\sum_{t=1}^{T-1}\ell_{\text{rec}}\!\left(\hat{I}^{\text{flow}}_{t},\, I_t\right),\quad
\mathcal{L}_{\text{trans}}
=
\frac{1}{T-1}\sum_{t=1}^{T-1}\ell_{\text{rec}}\!\left(\hat{I}^{\text{pix}}_{t},\, I_t\right).
\end{equation}
Here, $\ell_{\text{rec}}(\cdot,\cdot)$ denotes the $\ell_1$ reconstruction loss in our implementation.

\textbf{Segmentation loss $\mathcal{L}_{\text{seg}}$.}
Let $M^{\text{ED}}, M^{\text{ES}}\in\{0,\dots,C-1\}^{Z\times H\times W}$ denote sparse segmentation labels at the end-diastolic (ED) and end-systolic (ES) frames, respectively.
These auxiliary labels provide direct anatomical supervision at key phases of the cardiac cycle, encouraging the synthesized cine to preserve ventricular boundaries and myocardial thickness.
The Segmentation Branch predicts voxel-wise logits $S(\cdot)\in\mathbb{R}^{C\times Z\times H\times W}$, where $C=3$ corresponds to clinically relevant cardiac structures: the left ventricle (LV), the right ventricle (RV), and the left ventricular myocardium (MYO).

Segmentation is supervised at both ED and ES phases, on the ED input volume and the pixel-translated ES volume:
\begin{equation}
\mathcal{L}_{\text{seg}}
=
\lambda_{\text{seg}}\Big(
\ell_{\text{seg}}\!\left(S(I_0),\, M^{\text{ED}}\right)
+
\ell_{\text{seg}}\!\left(S(\hat{I}^{\text{pix}}_{T-1}),\, M^{\text{ES}}\right)
\Big),
\end{equation}
where $\ell_{\text{seg}}(\cdot,\cdot)$ is implemented as Dice+cross-entropy.

\textbf{Cross-domain consistency loss $\mathcal{L}_{\text{cons}}$.}
To couple the two synthesis domains (deformation-based vs. translation-based), we introduce a teacher--student cross-domain consistency constraint.
Specifically, the flow-guided branch serves as a teacher to regularize the pixel-translation branch:
\begin{equation}
\mathcal{L}_{\text{cons}}
=
\lambda_{\text{cons}}\,
\ell_{\text{con}}\!\left(\hat{I}^{\text{pix}},\, \text{stopgrad}(\hat{I}^{\text{flow}})\right).
\end{equation}
where $\ell_{\text{con}}$ denotes the Charbonnier penalty and $\text{stopgrad}(\cdot)$ blocks gradient propagation through the flow-guided branch.
This constraint encourages the appearance dynamics synthesized by the pixel-translation branch to follow the explicit motion guidance, reducing temporal drift across frames.

\textbf{Adversarial loss $\mathcal{L}_{\text{adv}}$.}
We further apply adversarial learning using two 3D PatchGAN discriminators with hinge loss to enhance realism and anatomical fidelity.
Both discriminators are conditioned on the ED reference frame by concatenating the input pair $[I_0,\cdot]$.
The flow discriminator $D_{\text{flow}}$ distinguishes real pairs $[I_0,I_t]$ from generated pairs $[I_0,\hat{I}^{\text{flow}}_t]$, while the pixel discriminator $D_{\text{pix}}$ distinguishes $[I_0,I_t]$ from $[I_0,\hat{I}^{\text{pix}}_t]$ for sampled time indices $t$.
We denote the corresponding generator adversarial losses as $\mathcal{L}_{\text{adv-flow}}$ and $\mathcal{L}_{\text{adv-pix}}$:
\begin{equation}
\mathcal{L}_{\text{adv}}
=
\lambda_{\text{adv-flow}}\mathcal{L}_{\text{adv-flow}}
+
\lambda_{\text{adv-pix}}\mathcal{L}_{\text{adv-pix}}.
\end{equation}

\textbf{Overall objective.}
The discriminator loss is defined as
\begin{equation}
\mathcal{L}_{D}
=
\mathcal{L}_{D}^{\text{flow}}
+
\mathcal{L}_{D}^{\text{pix}}.
\end{equation}
The generator is trained by minimizing a weighted sum of reconstruction,
segmentation, cross-domain consistency, and adversarial objectives:

\begin{equation}
\mathcal{L}_{G}
=
\mathcal{L}_{\text{recon}}
+
\mathcal{L}_{\text{seg}}
+
\mathcal{L}_{\text{cons}}
+
\mathcal{L}_{\text{adv}}.
\end{equation}
The weighting coefficients
$\lambda_{\text{flow-img}}, \lambda_{\text{pix-img}}, \lambda_{\text{seg}},
\lambda_{\text{cons}}, \lambda_{\text{adv-flow}}, \lambda_{\text{adv-pix}}$
are fixed as described in Sec.~\ref{sec:experiments}.

\section{Experiments}
\label{sec:experiments}
We conduct extensive experiments and ablation studies to evaluate the effectiveness of the proposed method, and compare it against representative GAN-based and diffusion-based approaches.

\begin{figure}[t]
    \centering
    \includegraphics[width=\textwidth]{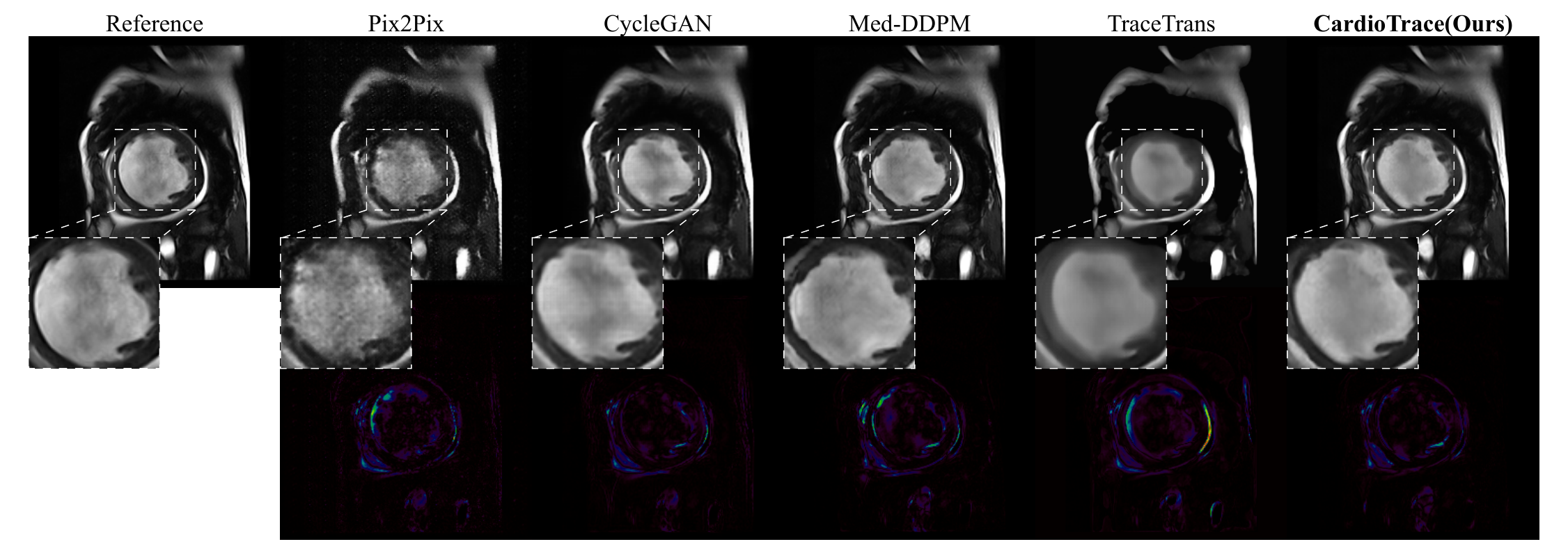}
    \caption{Qualitative comparison on the M\&M dataset.
    From left to right: ground-truth reference, Pix2Pix, CycleGAN, Med-DDPM, TraceTrans, and the proposed CardioTrace.
    For each method, the top row shows the synthesized end-systolic (ES) frame conditioned on the end-diastolic (ED) input, with dashed boxes indicating regions of interest that are enlarged below.
    The bottom row visualizes the corresponding absolute error maps with respect to the ground-truth ES frame.
    Compared with baseline methods, CardioTrace better preserves ventricular morphology and boundary sharpness, while reducing errors in regions undergoing large cardiac deformation.}
    \label{mnms_fig}
\end{figure}

\textbf{Datasets and Preprocessing.} 
Experiments are conducted on two publicly available cardiac cine MRI datasets, M\&M~\cite{9458279} and ACDC~\cite{8360453}, which contain 375 and 150 subjects, respectively.
For each subject, the full short-axis cine MRI volumes are used, resulting in 4D data with dimensions $(T, Z, H, W)$.
The temporal interval between end-diastole (ED) and end-systole (ES) is identified and uniformly sampled to obtain a fixed-length cine sequence of $T=8$ frames.
All samples are spatially resized to a resolution of $8 \times 10 \times 256 \times 256$.
Image intensities are normalized on a per-volume basis using min–max normalization.
Corresponding segmentation annotations at ED and ES are used for anatomical supervision.
The datasets are split into training, validation, and test sets with a ratio of 8:1:1 at the subject level.

\begin{figure}[htbp]
    \centering
    \includegraphics[width=\textwidth]{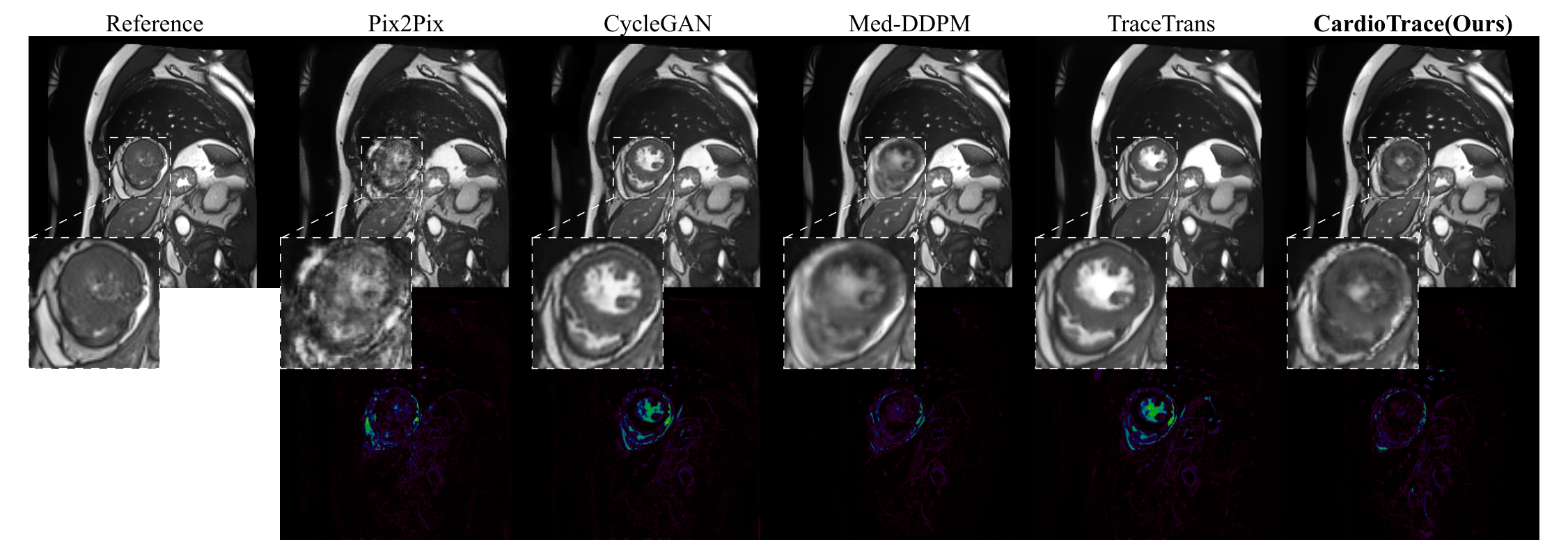}
    \caption{Qualitative comparison of CardioTrace with several baseline methods on the ACDC dataset.
    The visualization layout is identical to Fig.~\ref{mnms_fig}.
    CardioTrace produces end-systolic images that are closer to the ground truth, with fewer artifacts and reduced errors in regions undergoing large deformation, demonstrating improved anatomical consistency across the cardiac cycle.}
    \label{acdc_fig}
\end{figure}

\textbf{Implementation Details.}
Our model is implemented in PyTorch Lightning and trained for 200 epochs with a batch size of 8 on a single NVIDIA A100 (40GB). We use Adam with learning rates of $3 \times 10^{-4}$ and $6 \times 10^{-4}$ for the generator and discriminator, respectively, and the base channel width is 64. The training objective is a weighted sum of losses with fixed coefficients: flow-guided and pixel-translation reconstruction terms use $\lambda_{\text{flow-img}}=10.0$ and $\lambda_{\text{pix-img}}=20.0$, segmentation at ED/ES uses $\lambda_{\text{seg}}=0.5$, branch consistency uses $\lambda_{\text{cons}}=1.0$, and adversarial losses for the flow-guided and pixel-translation branches use $\lambda_{\text{adv-flow}}=\lambda_{\text{adv-pix}}=0.1$. For the M\&Ms dataset, meta-conditioning is enabled using categorical embeddings of dimension 16 and a fused meta feature dimension of 256, while meta-conditioning is disabled for ACDC.

\textbf{Evaluation Metrics.} We evaluate CardioTrace and competing baselines using four standard image similarity metrics: Structural Similarity Index (SSIM)~\cite{wang2004image}, Peak Signal-to-Noise Ratio (PSNR), Mean Absolute Error (MAE), and Normalized Mutual Information (NMI). Higher SSIM, PSNR, and NMI, and lower MAE indicate better synthesis quality.

\begin{table}[t]
    \centering
    \setlength{\tabcolsep}{2.2pt}
    \renewcommand{\arraystretch}{1.2}
    \caption{Quantitative comparison on the M\&Ms dataset. CardioTrace obtained the best result in the majority of metrics on this dataset.}
    \begin{tabular}{l|cccc}
        \hline
        \textbf{Method} & SSIM$(\%)$$\uparrow$  & MAE$\downarrow$   & PSNR$\uparrow$   & NMI$\uparrow$\\
        \hline
         Pix2Pix~\cite{isola2018imagetoimagetranslationconditionaladversarial} & $78.65\pm6.15$  & $\mathbf{0.0412\pm0.0131}$ & $20.89\pm2.78$ & $0.3527\pm0.0515$\\
         CycleGAN~\cite{zhu2020unpairedimagetoimagetranslationusing} & $78.57\pm9.55$ & $0.0445\pm0.0210$ & $21.15\pm3.19$ & $0.4016\pm0.0963$\\
         Med-DDPM~\cite{Dorjsembe_2024} & $71.5\pm0.88$ & $0.0492\pm0.0196$ & $21.91\pm1.42$ & $0.4087\pm0.0303$\\
         TraceTrans~\cite{luo2025tracetrans}& $67.15\pm6.77$ & $0.0497\pm0.0150$ & $22.08\pm2.21$ & $0.3326\pm0.0576$\\ 
         \textbf{CardioTrace}&$\mathbf{80.04 \pm 9.14}$ & $0.0472 \pm 0.0250$ & $\mathbf{22.34 \pm 3.31}$ & $\mathbf{0.4235 \pm 0.1045}$\\
        \hline
    \end{tabular}
    \label{mnms_result}
\end{table}

\textbf{Evaluation Performance on M\&M dataset.} 
Quantitative results on the M\&M dataset are reported in Table~\ref{mnms_result}. 
CardioTrace achieves the best performance on most metrics, indicating improved perceptual quality and structural consistency with the ground-truth cine frames. 
Qualitative results in Fig.~\ref{mnms_fig} show that CardioTrace generates end-systolic frames closer to the ground truth, with clearer myocardial boundaries and fewer motion artifacts. 
Error heatmaps further demonstrate reduced discrepancies in regions with large deformation, validating the benefit of explicit motion tracing.

\textbf{Evaluation Performance on ACDC dataset.}
Quantitative results on the ACDC dataset are reported in Table~\ref{acdc_result}. 
CardioTrace achieves the best overall performance on MAE, PSNR, and NMI, and remains competitive in SSIM, indicating improved reconstruction accuracy and structural consistency.
Qualitative comparisons in Fig.~\ref{acdc_fig} further show that CardioTrace produces end-systolic images closer to the ground truth, with fewer artifacts and reduced errors in regions undergoing large deformation.

\begin{table}[t]
    \centering
    \renewcommand{\arraystretch}{1.2}
    \caption{Quantitative comparison on the ACDC dataset. CardioTrace obtained the best result in the majority of metrics on this datasets.}
    \begin{tabular}{l|cccc}
        \hline
        \textbf{Method} & SSIM$(\%)$$\uparrow$  & MAE$\downarrow$   & PSNR$\uparrow$   & NMI$\uparrow$\\
        \hline
         Pix2Pix~\cite{isola2018imagetoimagetranslationconditionaladversarial} & $84.62\pm5.56$ & $0.0293\pm0.0092$ & $24.89\pm2.15$ & $0.4797\pm0.0463$\\
         CycleGAN~\cite{zhu2020unpairedimagetoimagetranslationusing} & $\mathbf{85.82\pm6.26}$ & $0.0303\pm0.0105$ & $24.26\pm2.40$ & $0.5068\pm0.0550$\\
         Med-DDPM~\cite{Dorjsembe_2024} & $83.11\pm9.41$ & $0.0327\pm0.0115$ & $24.65\pm2.15$ & $0.5376\pm0.0406$\\
         TraceTrans~\cite{luo2025tracetrans}& $79.62\pm6.76$ & $0.0346\pm0.0106$ & $23.84\pm2.23$ & $0.4918\pm0.0493$\\
         \textbf{CardioTrace}& $85.31\pm8.66$ &$\mathbf{0.0281\pm0.0058}$  & $\mathbf{25.56\pm1.14}$ & $\mathbf{0.5812\pm0.0472}$\\
        \hline
    \end{tabular}
    \label{acdc_result}
\end{table}
\textbf{Ablation Study.}
Table~\ref{tab3} reports ablation results on the M\&M dataset to analyze the contribution of each component in CardioTrace. Removing the flow or segmentation branch consistently degrades performance, demonstrating the importance of explicit motion tracing and anatomical supervision. Importantly, removing the meta-conditioning further reduces performance, highlighting that prompt-style information improves robustness under heterogeneous acquisition protocols and patient variations. The full CardioTrace model achieves the best overall results, validating the effectiveness of jointly integrating all components.

\begin{table}[t]
    \centering
    \renewcommand{\arraystretch}{1.2}
    \caption{Quantitative ablation study evaluating the contribution of different model components to correspondence performance on the M\&M dataset.}
    \begin{tabular}{l|cccc}
        \hline
        \textbf{Method} & SSIM$(\%)$$\uparrow$  & MAE$\downarrow$   & PSNR$\uparrow$   & NMI$\uparrow$\\
        \hline
         w/o seg branch& $79.72\pm3.22$ & $0.0582\pm0.0093$ & $20.43\pm1.05$ & $0.4128\pm0.0382$\\
         w/o flow branch& $74.31\pm0.68$ & $0.0493\pm0.0102$ & $21.32\pm2.08$ & $0.3901\pm0.0312$\\
         w/o meta& $77.01\pm1.40$ & $0.0532\pm0.0168$ & $22.19\pm5.19$ & $0.3719\pm0.0273$\\
         flow branch only& $79.24\pm6.19$ & $0.0479\pm0.0175$ & $21.70\pm1.32$ & $0.4108\pm0.0661$\\
         \textbf{CardioTrace}&$\mathbf{80.04 \pm 9.14}$ & $\mathbf{0.0472 \pm 0.0250}$ & $\mathbf{22.34 \pm 3.31}$ & $\mathbf{0.4235 \pm 0.1045}$\\
         \hline
    \end{tabular}

    \label{tab3}
\end{table}

\section{Conclusion}
In this work, we presented CardioTrace, a motion-aware framework for cardiac cine MRI synthesis that couples image generation with explicit motion tracing and anatomical supervision. 
Unlike prior cine synthesis methods that primarily focus on appearance realism, CardioTrace provides an interpretable motion representation that links synthesized frames through anatomically meaningful correspondences, promoting temporal consistency. 
Experiments on cardiac cine MRI datasets demonstrate superior translation quality and structural fidelity, highlighting the method’s potential for cardiac applications that require accurate modeling of cardiac anatomy and motion.

    



%
%
%
\bibliographystyle{splncs04}
\bibliography{ref}

@misc{isola2018imagetoimagetranslationconditionaladversarial,
      title={Image-to-Image Translation with Conditional Adversarial Networks}, 
      author={Phillip Isola and Jun-Yan Zhu and Tinghui Zhou and Alexei A. Efros},
      year={2018},
      eprint={1611.07004},
      archivePrefix={arXiv},
      primaryClass={cs.CV},
      url={https://arxiv.org/abs/1611.07004}, 
}

@misc{zhu2020unpairedimagetoimagetranslationusing,
      title={Unpaired Image-to-Image Translation using Cycle-Consistent Adversarial Networks}, 
      author={Jun-Yan Zhu and Taesung Park and Phillip Isola and Alexei A. Efros},
      year={2020},
      eprint={1703.10593},
      archivePrefix={arXiv},
      primaryClass={cs.CV},
      url={https://arxiv.org/abs/1703.10593}, 
}

@article{Dorjsembe_2024,
   title={Conditional Diffusion Models for Semantic 3D Brain MRI Synthesis},
   volume={28},
   ISSN={2168-2208},
   url={http://dx.doi.org/10.1109/JBHI.2024.3385504},
   DOI={10.1109/jbhi.2024.3385504},
   number={7},
   journal={IEEE Journal of Biomedical and Health Informatics},
   publisher={Institute of Electrical and Electronics Engineers (IEEE)},
   author={Dorjsembe, Zolnamar and Pao, Hsing-Kuo and Odonchimed, Sodtavilan and Xiao, Furen},
   year={2024},
   month=jul, pages={4084–4093} }

@ARTICLE{9458279,
  author={Campello, Víctor M. and Gkontra, Polyxeni and Izquierdo, Cristian and Martín-Isla, Carlos and Sojoudi, Alireza and Full, Peter M. and Maier-Hein, Klaus and Zhang, Yao and He, Zhiqiang and Ma, Jun and Parreño, Mario and Albiol, Alberto and Kong, Fanwei and Shadden, Shawn C. and Acero, Jorge Corral and Sundaresan, Vaanathi and Saber, Mina and Elattar, Mustafa and Li, Hongwei and Menze, Bjoern and Khader, Firas and Haarburger, Christoph and Scannell, Cian M. and Veta, Mitko and Carscadden, Adam and Punithakumar, Kumaradevan and Liu, Xiao and Tsaftaris, Sotirios A. and Huang, Xiaoqiong and Yang, Xin and Li, Lei and Zhuang, Xiahai and Viladés, David and Descalzo, Martín L. and Guala, Andrea and Mura, Lucia La and Friedrich, Matthias G. and Garg, Ria and Lebel, Julie and Henriques, Filipe and Karakas, Mahir and Çavuş, Ersin and Petersen, Steffen E. and Escalera, Sergio and Seguí, Santi and Rodríguez-Palomares, José F. and Lekadir, Karim},
  journal={IEEE Transactions on Medical Imaging}, 
  title={Multi-Centre, Multi-Vendor and Multi-Disease Cardiac Segmentation: The M\&Ms Challenge}, 
  year={2021},
  volume={40},
  number={12},
  pages={3543-3554},
  doi={10.1109/TMI.2021.3090082}}

@ARTICLE{8360453,
  author={Bernard, Olivier and Lalande, Alain and Zotti, Clement and Cervenansky, Frederick and Yang, Xin and Heng, Pheng-Ann and Cetin, Irem and Lekadir, Karim and Camara, Oscar and Gonzalez Ballester, Miguel Angel and Sanroma, Gerard and Napel, Sandy and Petersen, Steffen and Tziritas, Georgios and Grinias, Elias and Khened, Mahendra and Kollerathu, Varghese Alex and Krishnamurthi, Ganapathy and Rohé, Marc-Michel and Pennec, Xavier and Sermesant, Maxime and Isensee, Fabian and Jäger, Paul and Maier-Hein, Klaus H. and Full, Peter M. and Wolf, Ivo and Engelhardt, Sandy and Baumgartner, Christian F. and Koch, Lisa M. and Wolterink, Jelmer M. and Išgum, Ivana and Jang, Yeonggul and Hong, Yoonmi and Patravali, Jay and Jain, Shubham and Humbert, Olivier and Jodoin, Pierre-Marc},
  journal={IEEE Transactions on Medical Imaging}, 
  title={Deep Learning Techniques for Automatic MRI Cardiac Multi-Structures Segmentation and Diagnosis: Is the Problem Solved?}, 
  year={2018},
  volume={37},
  number={11},
  pages={2514-2525},
  doi={10.1109/TMI.2018.2837502}}

@article{luo2025tracetrans,
  title={TraceTrans: Translation and Spatial Tracing for Surgical Prediction},
  author={Luo, Xiyu and Li, Haodong and Cheng, Xinxing and Zhao, He and Hu, Yang and Song, Xuan and Zhang, Tianyang},
  journal={arXiv preprint arXiv:2510.22379},
  year={2025}
}

@article{zheng2024deformation,
  title={Deformation-recovery diffusion model (DRDM): instance deformation for image manipulation and synthesis},
  author={Zheng, Jian-Qing and Mo, Yuanhan and Sun, Yang and Li, Jiahua and Wu, Fuping and Wang, Ziyang and Vincent, Tonia and Papie{\.z}, Bart{\l}omiej W},
  journal={arXiv preprint arXiv:2407.07295},
  year={2024}
}

@misc{huang2024subspaceimplicitneuralrepresentations,
      title={Subspace Implicit Neural Representations for Real-Time Cardiac Cine MR Imaging}, 
      author={Wenqi Huang and Veronika Spieker and Siying Xu and Gastao Cruz and Claudia Prieto and Julia Schnabel and Kerstin Hammernik and Thomas Kuestner and Daniel Rueckert},
      year={2024},
      eprint={2412.12742},
      archivePrefix={arXiv},
      primaryClass={eess.IV},
      url={https://arxiv.org/abs/2412.12742}, 
}

@misc{qiu2024spatiotemporaldiffusionmodelpaired,
      title={Spatiotemporal Diffusion Model with Paired Sampling for Accelerated Cardiac Cine MRI}, 
      author={Shihan Qiu and Shaoyan Pan and Yikang Liu and Lin Zhao and Jian Xu and Qi Liu and Terrence Chen and Eric Z. Chen and Xiao Chen and Shanhui Sun},
      year={2024},
      eprint={2403.08758},
      archivePrefix={arXiv},
      primaryClass={eess.IV},
      url={https://arxiv.org/abs/2403.08758}, 
}

@inproceedings{vukadinovic2023gancmri,
  title={GANcMRI: Cardiac magnetic resonance video generation and physiologic guidance using latent space prompting},
  author={Vukadinovic, Milos and Kwan, Alan C and Li, Debiao and Ouyang, David},
  booktitle={Machine Learning for Health (ML4H)},
  pages={594--606},
  year={2023},
  organization={PMLR}
}

@article{wang2004image,
  title={Image quality assessment: from error visibility to structural similarity},
  author={Wang, Zhou and Bovik, Alan C and Sheikh, Hamid R and Simoncelli, Eero P},
  journal={IEEE transactions on image processing},
  volume={13},
  number={4},
  pages={600--612},
  year={2004},
  publisher={IEEE}
}

@article{yoo2020generative,
  title={A generative adversarial network approach to predicting postoperative appearance after orbital decompression surgery for thyroid eye disease},
  author={Yoo, Tae Keun and Choi, Joon Yul and Kim, Hong Kyu},
  journal={Computers in biology and medicine},
  volume={118},
  pages={103628},
  year={2020},
  publisher={Elsevier}
}

@inproceedings{cciccek20163d,
  title={3D U-Net: learning dense volumetric segmentation from sparse annotation},
  author={{\c{C}}i{\c{c}}ek, {\"O}zg{\"u}n and Abdulkadir, Ahmed and Lienkamp, Soeren S and Brox, Thomas and Ronneberger, Olaf},
  booktitle={International conference on medical image computing and computer-assisted intervention},
  pages={424--432},
  year={2016},
  organization={Springer}
}

@inproceedings{perez2018film,
  title={Film: Visual reasoning with a general conditioning layer},
  author={Perez, Ethan and Strub, Florian and De Vries, Harm and Dumoulin, Vincent and Courville, Aaron},
  booktitle={Proceedings of the AAAI conference on artificial intelligence},
  volume={32},
  number={1},
  year={2018}
}

@inproceedings{isola2017image,
  title={Image-to-image translation with conditional adversarial networks},
  author={Isola, Phillip and Zhu, Jun-Yan and Zhou, Tinghui and Efros, Alexei A},
  booktitle={Proceedings of the IEEE conference on computer vision and pattern recognition},
  pages={1125--1134},
  year={2017}
}
\end{document}